\documentclass[11pt,a4paper]{article}
\usepackage[T1]{fontenc}
\usepackage[utf8]{inputenc}
\usepackage[a4paper,left=24mm,right=24mm,top=23mm,bottom=23mm]{geometry}
\usepackage{amsmath,amssymb}
\usepackage{newtxtext,newtxmath}
\usepackage{graphicx}
\usepackage{booktabs,array}
\usepackage{microtype}
\usepackage{cite}
\usepackage[font=small,labelfont=bf,labelsep=period]{caption}
\usepackage{placeins}
\usepackage[hidelinks,unicode]{hyperref}
\makeatletter
\newcommand{\bstctlcite}[1]{\immediate\write\@auxout{\string\citation{#1}}}
\makeatother
\hypersetup{
  pdftitle={ISW-Agent: Integrating LLM-Driven Dialogue with n8n Workflow for Internal Solitary Wave Remote Sensing Image Processing},
  pdfauthor={Hao Zhang; Hengrui Liu; Lina Sun; Junmin Meng},
  pdfsubject={Internal solitary wave remote sensing and scientific workflow orchestration},
  pdfkeywords={internal solitary waves, remote sensing, LLM, n8n}
}
\title{\bfseries\LARGE ISW-Agent: Integrating LLM-Driven Dialogue with n8n Workflow for Internal Solitary Wave Remote Sensing Image Processing}
\author{Hao Zhang\thanks{Corresponding author: \href{mailto:zhanghao@fio.org.cn}{zhanghao@fio.org.cn}.},
  Hengrui Liu, Lina Sun, and Junmin Meng\\[5pt]
  \small First Institute of Oceanography, Ministry of Natural Resources,\\
  \small Qingdao, China}
\date{}
\begin{document}
\bstctlcite{bstcontrol}
\maketitle
\begin{abstract}
Internal solitary wave (ISW) remote sensing requires task parameters and spatial information to be transferred across image acquisition, detection, stripe extraction, and propagation prediction. Here, we present ISW-Agent, which integrates large language model (LLM) interpretation with n8n workflows and independent scientific models through structured task constraints, geospatial transformations, and stage-specific output validation. In tests of 120 Chinese instructions per backend, GPT-5.6 Sol and DeepSeek both achieved a strict task-match rate of 94.2\%, with all intents and execution decisions matching the reference. Sentinel-1 cases demonstrated continuous processing from dialogue requests to wavefront prediction for the specified scene. Of 50 timed scenes, 39 completed detection, extraction, and prediction with a median turnaround time of 48.04 s; 11 stopped after empty extraction. Across 3,408 controlled fault-injection runs, retries recovered 234 transient-failure conditions and output validation rejected 234 structural defects, while 61 provenance or completeness errors indicated the need for content-consistency checks. Overall, ISW-Agent thus connects natural-language requests to traceable spatial processing chains for multimodel ISW analysis.
\end{abstract}

\section{Introduction}

Internal solitary waves (ISWs) are nonlinear waves in stratified oceans whose propagation and breaking contribute to energy redistribution and mixing \cite{ref01}, \cite{ref02}. Synthetic aperture radar (SAR) records their surface signatures as bright and dark bands produced by the modulation of short-wave roughness. Research on these signatures has developed along three related directions: detection, stripe extraction, and propagation prediction \cite{ref03}, \cite{ref04}. Connecting these stages converts wave-packet detections in individual scenes into spatially and temporally referenced analysis products.

During the past two decades, satellite-based ISW analysis has advanced from wavelet detection \cite{ref05} to deep learning models such as Faster R-CNN \cite{ref06} and IWE-Net \cite{ref07}. Stripe extraction have progressed from edge-based and SVM methods \cite{ref08} to U-Net \cite{ref09}, \cite{ref10}. Propagation prediction has likewise developed from KdV-based solvers to machine learning models that combine satellite observations with forecasts \cite{ref11,ref12,ref13,ref14}. These methods are typically developed for specific inputs and evaluation objectives. Transferring images, detection boxes, stripe masks, and crestline coordinates between them still requires additional data preparation and parameter configuration.

This fragmentation creates two practical bottlenecks. Users must transfer intermediate products between programs and interpret the fields, coordinates, and file formats required by each service. In addition, a successful service response does not ensure that its output meets the requirements of the next stage. A truncated stripe contour, for example, may remain writable as a vector file while altering the spatial extent used to construct the initial wavefront. A reproducible processing chain therefore requires explicit task specification, compatible interfaces, and output validation.

Existing workflow platforms have been used in geoscience \cite{ref15}, but generally provide limited support for natural-language task specification and domain-specific decision rules. LLM-based agents offer a natural-language interface to geoscience and remote sensing tools \cite{ref16,ref17,ref18,ref19,ref20,ref21}. RS-Agent combines task reasoning with tool use \cite{ref18}, while ThinkGeo evaluates agents performing remote sensing operations \cite{ref20}. For ISW analysis, task interpretation must also support continuous transfer of spatial products: image and crestline identities must remain consistent across models, and service failures require explicit handling rules. We therefore propose ISW-Agent, which separates natural-language interpretation from deterministic execution. The architecture is evaluated through dialogue contracts, preservation of spatial representations, execution time, and terminal outcomes under controlled faults.

\section{System Design}

\begin{figure}[!htbp]
\centering
\includegraphics[width=\linewidth]{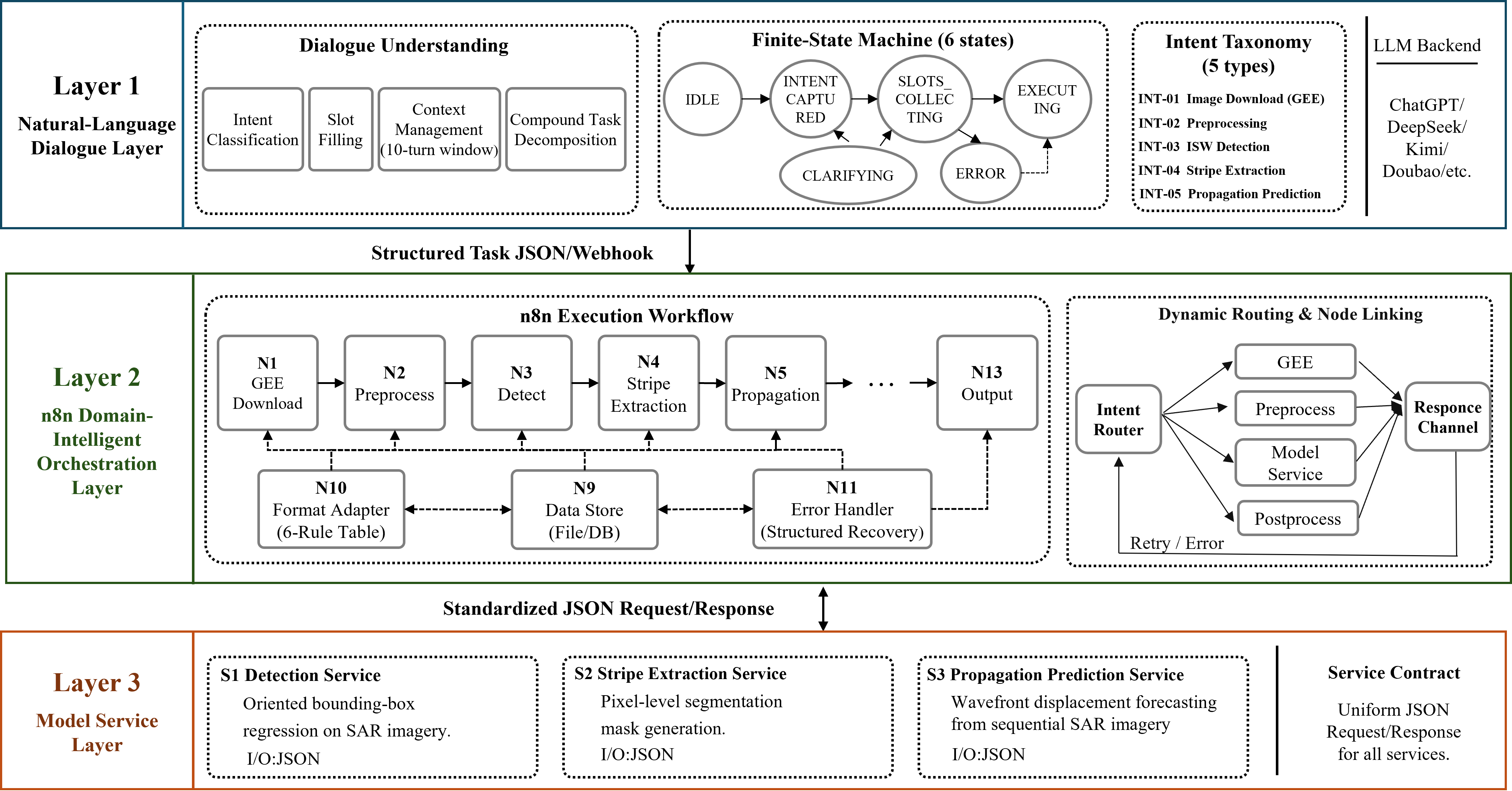}
\caption{Three-layer architecture of ISW-Agent: dialogue understanding, n8n-based domain-aware orchestration, and scientific model services. Nodes represent logical modules.}
\label{fig:01}
\end{figure}

ISW-Agent comprises dialogue understanding, domain-aware orchestration, and model services, which handle task specification, execution constraints, and scientific computation, respectively. Figure~\ref{fig:01} shows the data flow between these layers.

Natural-language requests are first converted into explicit operations and target objects. The n8n routing layer and a shared controller then check dependencies, acquire ISW SAR imagery through Google Earth Engine (GEE), and invoke detection, stripe-extraction, and propagation-prediction models. The n8n workflow uses a Webhook entry point to receive requests and HTTP request nodes to call the shared controller, which centralizes parameter validation, retries, and output checks. This design constrains LLM outputs through deterministic rules and allows different invocation routes to share the same scientific computations.

\subsection{Dialogue Understanding Layer}

The dialogue layer supports OpenAI-compatible LLM backends, including ChatGPT, DeepSeek, Kimi, and Doubao. It converts user intentions into processing tasks with explicit targets and complete parameters. Each request is mapped to a task contract \(\tau\) = (I, S, D, P, O), where I denotes ordered intents, S spatiotemporal slots, D sensor and data specifications, P model parameters, and O output and provenance requirements. The LLM proposes field values, while the controller supplies D from the asset registry and P and O from fixed service configurations. The task catalogue comprises GEE-based SAR image acquisition, preprocessing, ISW detection, stripe extraction, and propagation prediction. For a selected SAR image, the system identifies the required operations, image identity, target stripe, and forecast interval. These fields define the computational target and scope; the data inventory and fixed configurations supply the spatial reference, model weights, and output conventions. Flexible natural-language expressions are thus converted into verifiable task constraints.

All LLM backends use the same Chinese prompt, which specifies operation dependencies, missing-parameter handling, and context inheritance. Detection and stripe extraction are selected independently according to the request. Prediction requires a crestline extracted in the same run, without automatically adding detection. If the image, stripe identifier, or forecast interval is missing, the system retains known information and requests clarification. Subsequent instructions may inherit an established image or replace previous operations. Tasks are classified as executable, requiring clarification, or rejected. Complete requests with valid dependencies proceed to computation; ambiguous targets, out-of-range parameters, and conflicting information trigger clarification; out-of-domain or unsupported requests are rejected. These rules identify dialogue errors before spatial processing and keep missing information explicit.

\begin{table}[!htbp]
\centering
\caption{Task contracts for the five ISW-Agent intents.}
\label{tab:01}
\footnotesize
\setlength{\tabcolsep}{2.3pt}
\renewcommand{\arraystretch}{1.18}
\begin{tabular}{@{}>{\raggedright\arraybackslash}p{0.075\linewidth}>{\raggedright\arraybackslash}p{0.15\linewidth}>{\raggedright\arraybackslash}p{0.23\linewidth}>{\raggedright\arraybackslash}p{0.29\linewidth}>{\raggedright\arraybackslash}p{0.14\linewidth}@{}}
\toprule
\textbf{Intent} & \textbf{Operation} & \textbf{Required fields} & \textbf{Representative optional fields} & \textbf{Primary output} \\
\midrule
INT-01 & Image download & sensor, region, time\_start, time\_end & Satellite, product type, polarization, orbit direction & Raw image paths \\
\addlinespace[4pt]
INT-02 & Preprocessing & Input source & Resolution, speckle filter, land mask, output CRS, clipping region & Preprocessed image paths \\
\addlinespace[4pt]
INT-03 & ISW detection & Input source & Confidence threshold, NMS IoU threshold, model variant, output format & Detection results \\
\addlinespace[4pt]
INT-04 & Stripe extraction & Input image & Model variant, output format & Segmentation mask \\
\addlinespace[4pt]
INT-05 & Propagation prediction & Stripe data, target identifier, forecast interval & Environmental data, output format & Predicted trajectories \\
\bottomrule
\end{tabular}
\end{table}

\subsection{Domain-Aware Orchestration Layer}

Domain-aware orchestration connects independent models through data dependencies. GEE imagery, acquisition times, and spatial references provide a common input basis, while extracted ISW stripes supply the initial conditions for prediction in the same run. Each stage proceeds only when its inputs are available and the preceding outputs pass validation. Task records link input identities, model configurations, and output files. Continuity therefore depends on whether intermediate spatial information meets downstream requirements. Complementary checks address file readability, coordinates, and geometry. Input images must have an invertible affine transform and a WGS84 spatial reference. Detection counts must be internally consistent, and the number of extracted features must match the service response. An empty extraction is a valid output but provides no crestline for prediction. Applying these checks to output files identifies truncation, invalid geometry, and count mismatches at model interfaces.

Fault handling distinguishes recoverable service interruptions from invalid outputs. Rate limits, service unavailability, and transport timeouts trigger up to three attempts at the current stage, with increasing waiting intervals. Unparseable responses or invalid spatial structures terminate further processing while preserving completed outputs. Each scientific service call has a 300 s deadline. Bounded retries recover transient interruptions, while output validation maintains consistency between coupled models.

\subsection{Model Service Encapsulation}

The model service layer links SAR imagery, stripe geometry, and environmental features through independent detection, extraction, and prediction units. Requests specify input data, model configurations, and output locations. A task identifier links the data and configuration of each call, and the extracted vector file feeds prediction within the same run. Detection and stripe extraction share single-class YOLO11n-seg network, using multiscale feature fusion to generate bounding boxes and binary masks. Both operate on the full scene, with segmentation independent of box-based cropping to preserve stripe continuity across the observed area. The YOLO11n-seg network combines Conv, C3k2, SPPF, and C2PSA modules and fuses features at three scales for detection and segmentation. The inference image-size parameter is 1024 pixels, with confidence and nonmaximum suppression (NMS) intersection-over-union (IoU) thresholds of 0.07 and 0.30, respectively.

Stripe masks are georeferenced and converted to polygons, from which the selected initial wavefront is constructed. For multipart geometries, the largest component is retained. A centerline is obtained by skeletonization and main-path extraction in a local planar coordinate system. After standardizing endpoint order, it is resampled to 200 points along the crestline. This geometric wavefront represents the center of the SAR-visible stripe, preserves its curvature, and provides ordered spatial inputs for pointwise propagation calculations. Each wavefront point is represented by five features: longitude, latitude, water depth, and upper- and lower-layer density descriptors p\(_1\) and p\(_2\). A two-layer ConvLSTM with 64 hidden channels models relationships between adjacent points \cite{ref13}. The sequence represents spatial positions along the wavefront, and a target decoder converts the outputs to propagation speed and azimuth. For a specified forecast interval, spherical displacement is computed from the predicted speed and direction:

\begin{equation}
\label{eq:displacement}
p_i(t+\Delta t)=G\bigl[p_i(t),\,v_i\Delta t,\,\theta_i\bigr].
\end{equation}

Here, \(v_i\) and \(\theta_i\) are the propagation speed and azimuth at point i, respectively, with azimuth measured clockwise from geographic north. The forecast interval is \(\Delta t\), and \(G\)(\(\cdot\)) denotes the forward geographic-displacement operator. Postprocessing retains finite-valued predictions and applies a 10 km coastal exclusion and a 40 km distance constraint relative to the reference point set. A weighted spline fitted to the valid points yields the predicted wavefront.

\section{Results and Discussion}

\subsection{Image Acquisition and Model Data}

The dataset comprises 50 Sentinel-1 scenes acquired from 19 February 2022 to 17 May 2026 over the Maluku Sea, near 125\(^{\circ}\)--127.0\(^{\circ}\)E and 1\(^{\circ}\)S--2\(^{\circ}\)N. All scenes use interferometric wide-swath (IW) mode and VV polarization. Processing uses SAR quicklooks at a nominal display scale of 100 m. Linear VV intensity is converted to decibels and mapped to grayscale over \(-\)25 to 0 dB. This representation supports evaluation of image processing and spatial information transfer. All 50 scenes are included in the workflow timing experiment, and their outputs are also used in fault testing. Environmental inputs for propagation prediction are derived from GEBCO 2024 bathymetry and WOA2023 monthly temperature and salinity. Each wavefront point is matched to topography and the density descriptors for the corresponding month, linking observed geometry to local environmental conditions.

\begin{figure}[!htbp]
\centering
\includegraphics[width=\linewidth]{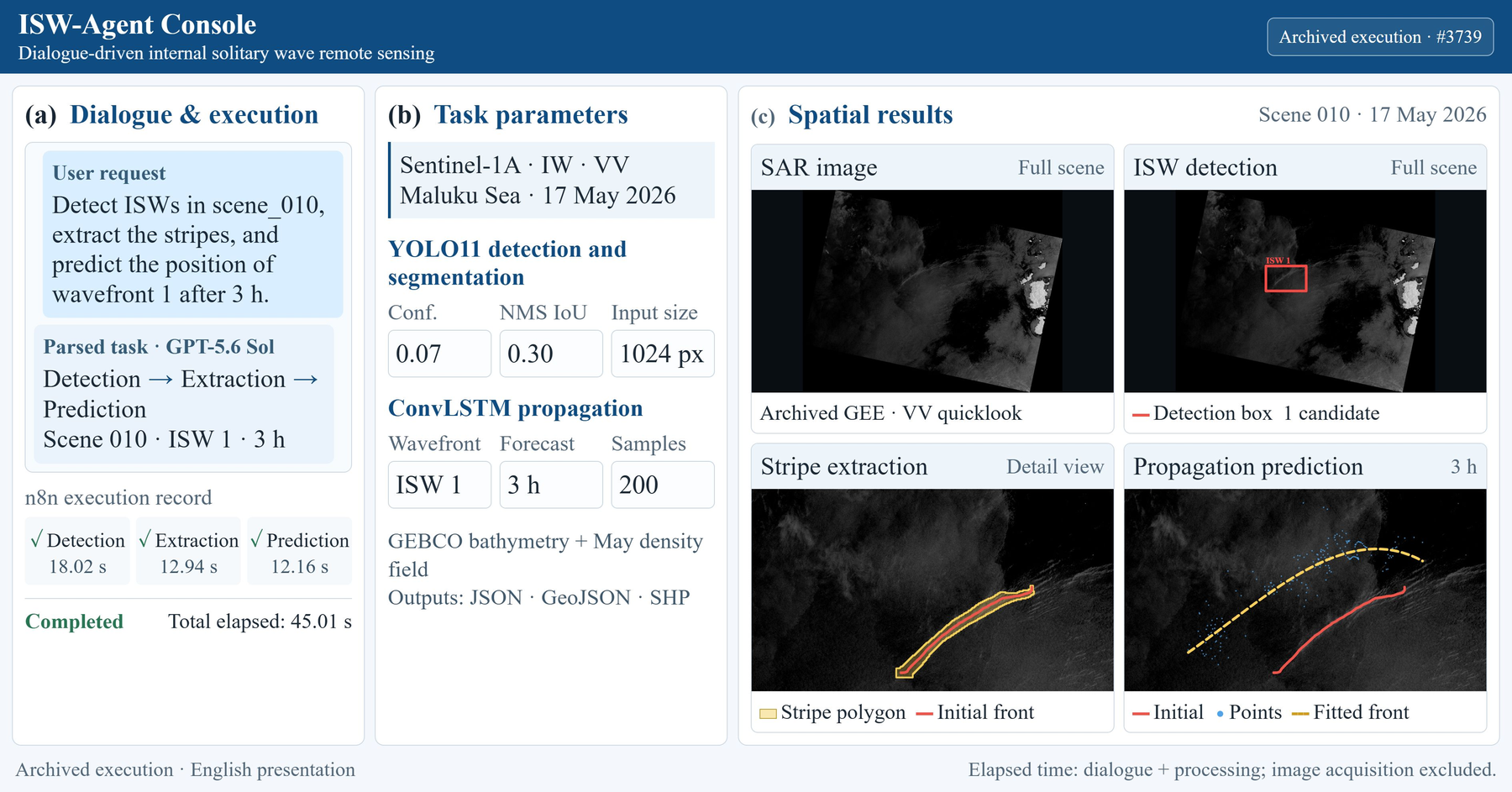}
\caption{Simplified ISW-Agent interface and processing of an archived GEE image. (a) Natural-language interpretation and n8n execution status. (b) Image information and key processing parameters. (c) The Sentinel-1A SAR quicklook acquired over the Maluku Sea on 17 May 2026 and archived after GEE download, with the corresponding ISW detection, stripe extraction, and 3 h propagation prediction.}
\label{fig:02}
\end{figure}

\subsection{Dialogue Understanding and Execution Decisions}

The dialogue dataset contains 120 Chinese instructions, with 10 examples in each of 12 task categories: detection only, extraction only, compound processing, prediction, three types of missing required information, out-of-domain requests, contextual references and updates, invalid forecast intervals, and unknown images. Examples vary the image identifier, stripe number, forecast interval, and colloquial prefix. Twenty context-dependent examples include an explicit preceding task to test object inheritance and operation replacement. Reference answers are generated from explicit inputs and predefined execution rules. All backends receive the same examples and prompt.

We report intent accuracy, microaveraged slot F1, and strict task-match rate. Intent accuracy is defined as \(\mathrm{Acc}_{\mathrm{intent}}=\frac{1}{N}\sum_{i=1}^{N}\mathbb{I}(\hat{y}_i=y_i),\) where \(y_i\) and \(\hat{y}_i\) denote the reference and predicted intents, respectively. Slot precision, recall, and F1 are computed by microaveraging, with each normalized slot key--value pair treated as a matching unit:

\begin{equation}
\label{eq:slot}
P_{\mathrm{slot}}=\frac{\sum\mathrm{TP}}{\sum\mathrm{TP}+\sum\mathrm{FP}},
\qquad R_{\mathrm{slot}}=\frac{\sum\mathrm{TP}}{\sum\mathrm{TP}+\sum\mathrm{FN}},
\qquad F1_{\mathrm{slot}}=\frac{2P_{\mathrm{slot}}R_{\mathrm{slot}}}{P_{\mathrm{slot}}+R_{\mathrm{slot}}}.
\end{equation}

GPT-5.6 Sol, DeepSeek V4 Pro, Kimi K2.6, and Doubao Seed 2.1 each process the 120 instructions once, without automatic retries for failed requests. Evaluation distinguishes operation intents, parameter slots, complete tasks, and execution decisions. Intent accuracy compares operation combinations; slot F1 aggregates matches for nonempty image, stripe, and forecast-interval values; strict matching requires agreement on all task information. Metrics over all requests retain cases without a valid response, and valid-response counts are reported separately to distinguish semantic interpretation from interface availability.

\begin{table}[!htbp]
\centering
\caption{Results for 120 requests per backend. Strict matches and correct decisions are counts. Slot F1 includes all requests. Latency is the median for valid responses.}
\label{tab:02}
\small
\setlength{\tabcolsep}{9pt}
\renewcommand{\arraystretch}{1.18}
\begin{tabular}{@{}lccccc@{}}
\toprule
\textbf{Backend} & \textbf{\shortstack{Valid\\responses}} & \textbf{\shortstack{Strict\\matches}} & \textbf{\shortstack{Correct\\decisions}} & \textbf{Slot F1} & \textbf{\shortstack{Latency\\(s)}} \\
\midrule
GPT-5.6 Sol & 120 & 113 & 120 & 0.978 & 1.73 \\
DeepSeek & 120 & 113 & 120 & 0.978 & 2.93 \\
Kimi & 27 & 26 & 26 & 0.307 & 7.90 \\
Doubao & 95 & 95 & 95 & 0.815 & 15.70 \\
\bottomrule
\end{tabular}
\end{table}

\begin{figure}[!htbp]
\centering
\includegraphics[width=\linewidth]{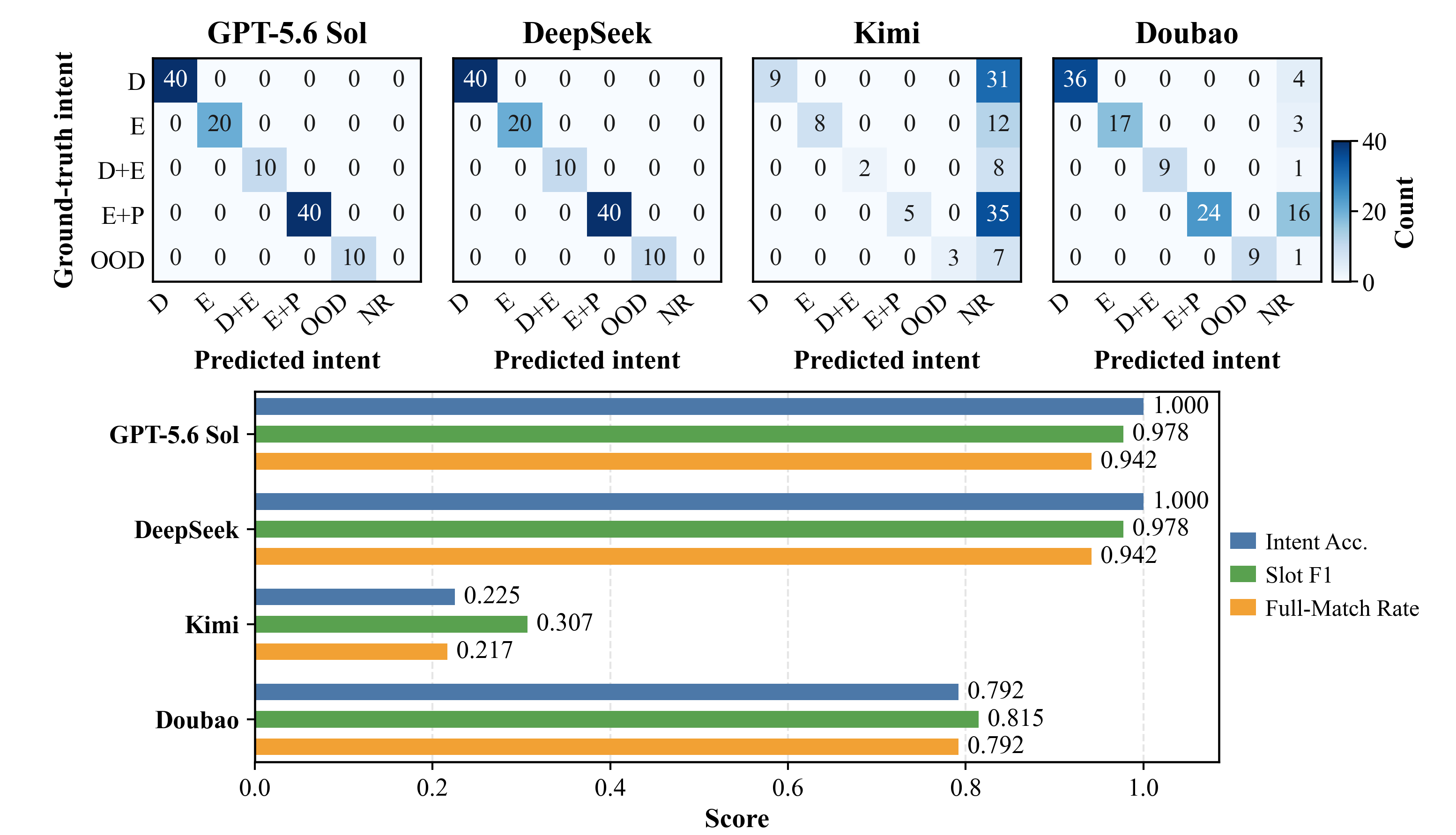}
\caption{Dialogue-understanding results for 120 Chinese instructions. The upper panels show intent confusion matrices: D, detection; E, extraction; P, prediction; OOD, out-of-domain request; and NR, no valid response. The lower panels show intent accuracy, slot F1, and task-match rate.}
\label{fig:03}
\end{figure}

GPT-5.6 Sol and DeepSeek correctly identified operation intents and execution decisions for all 120 requests, each achieving a strict task-match rate of 94.2\% and a slot F1 of 0.978 (Fig.~\ref{fig:03}). Each backend set seven invalid forecast intervals to null, whereas the reference retained the original values. All seven cases correctly triggered clarification, so the mismatch reflects ambiguity in parameter-representation rules without producing out-of-range predictions. Both backends also handled all 20 examples with a specified preceding task correctly, supporting object inheritance and task updates under explicit context. Kimi and Doubao returned 27 and 95 valid responses, with strict matching in 26/27 and 95/95, respectively. Their lower scores over all requests mainly reflect 93 rate-limit errors, 13 access denials, and 12 timeouts. Figure~\ref{fig:03} reports requests without valid responses separately from intent classes. Executable dialogue tasks thus depend on both parameter interpretation and service availability. Deterministic validation did not alter execution decisions for the valid outputs in this test; it made the conditions for downstream computation explicit.

\subsection{End-to-End ISW Processing and Propagation Prediction}

\begin{figure}[!htbp]
\centering
\includegraphics[width=\linewidth]{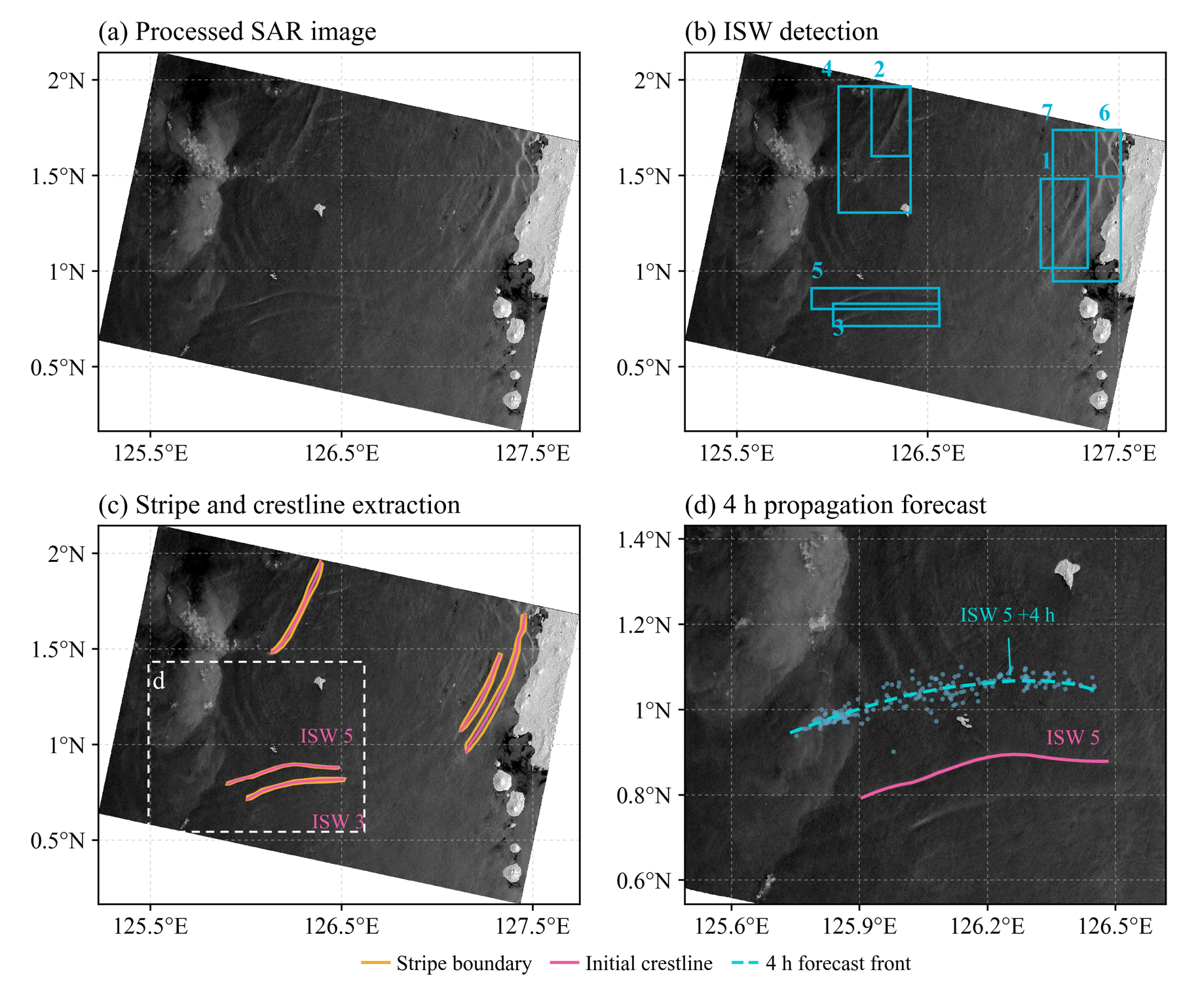}
\caption{ISW processing for a Sentinel-1A SAR image acquired over the Maluku Sea at 21:17 UTC on 6 March 2026. (a) Processed SAR image. (b) ISW detections, labeled by target identifier. (c) Stripe polygon boundaries and initial crestlines. (d) The initial ISW 5 crestline and its 4 h forecast front.}
\label{fig:04}
\end{figure}

The following case studies illustrate the spatial conversion from SAR signatures to stripe geometries and forecast fronts. Fig.~\ref{fig:04} uses a Sentinel-1A SAR image acquired over the Maluku Sea at 21:17 UTC on 6 March 2026. In the workflow, GEE retrieves imagery by sensor, acquisition time, and region; GPT-5.6 Sol interprets the target and forecast interval; and n8n links services according to their dependencies. The georeferenced image in Fig.~\ref{fig:04}a provides a common spatial basis for subsequent outputs. Fig.~\ref{fig:04}b contains seven candidate boxes covering curved ISW signatures with different orientations and scales. Fig.~\ref{fig:04}c shows polygon boundaries and central crestlines for five stripes, refining the candidate regions into geometries with defined extents and orientations. The initial crestline of the selected target, ISW 5, is approximately 65.7 km long and is resampled to 200 points. Fig.~\ref{fig:04}d presents the archived 4 h forecast curve, gives a mean speed of 1.71 m s\(^{-1}\), with pointwise speeds of 0.58--2.75 m s\(^{-1}\) and corresponding 4 h displacements of approximately 8.3--39.6 km.

\begin{figure}[!htbp]
\centering
\includegraphics[width=\linewidth]{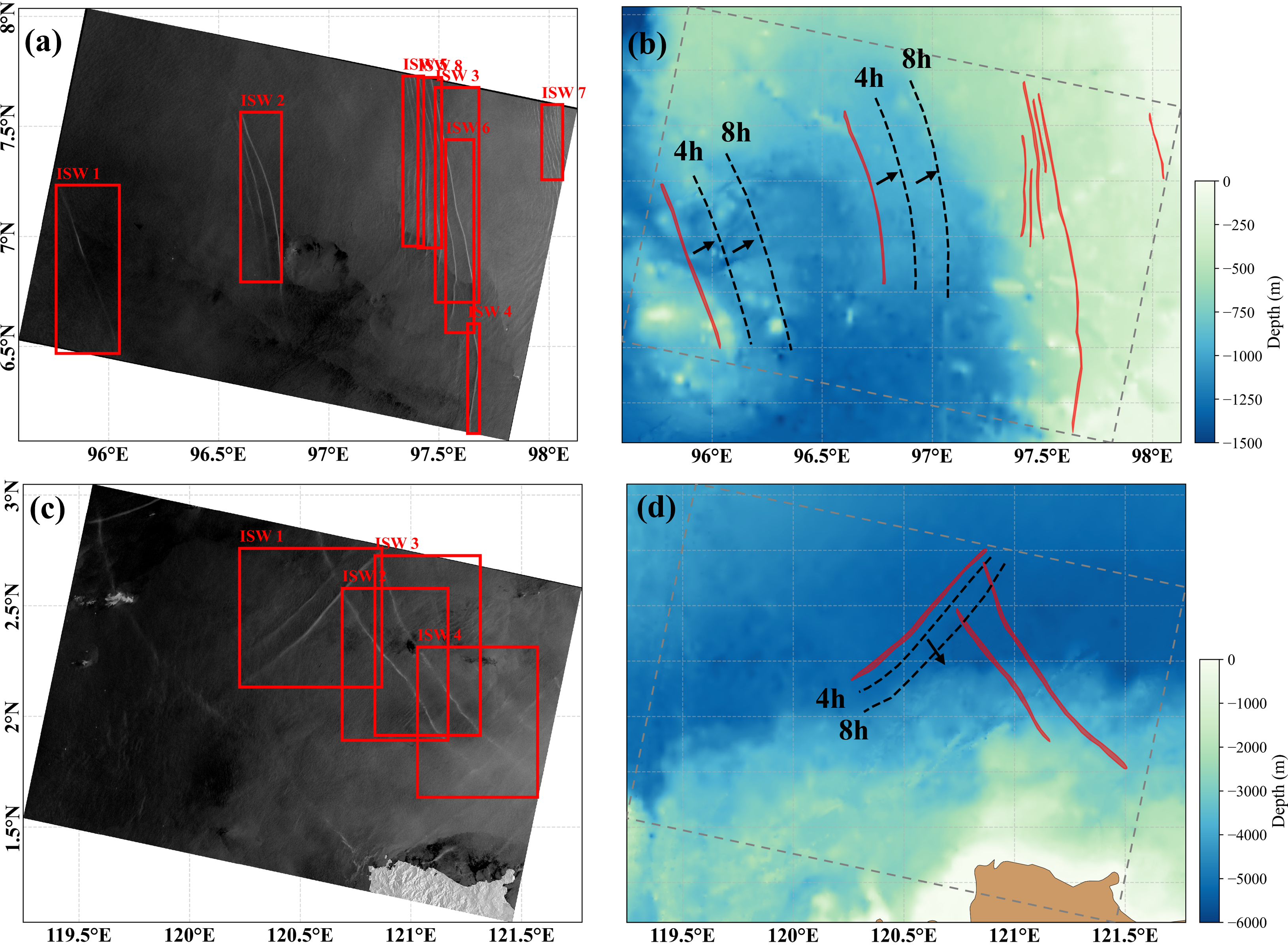}
\caption{ISW detection, stripe extraction, and propagation prediction in the Andaman and Celebes Seas. (a) Sentinel-1A SAR image and detections in the Andaman Sea on 23 December 2019. (b) Corresponding extraction results and 4 h and 8 h forecast fronts for ISW 1 and ISW 2. (c) Sentinel-1A SAR image and detections in the Celebes Sea on 22 August 2024. (d) Corresponding extraction results and 4 h and 8 h forecast fronts for ISW 1.}
\label{fig:05}
\end{figure}

Additional Andaman and Celebes Sea cases yielded eight and four candidates under the same detection settings. In the Andaman Sea scene acquired on 23 December 2019, stripes extend approximately north--south and are more closely spaced in the eastern part of the image (Fig.~\ref{fig:05}a). For selected ISW 1 and ISW 2, the mean speeds derived from initial-to-predicted wavefront displacement are 2.11 and 2.21 m s\(^{-1}\) over 0--4 h and 2.20 and 2.18 m s\(^{-1}\) over 0--8 h, respectively. The Celebes Sea scene acquired on 22 August 2024 contains several groups of curved stripes with different orientations (Fig.~\ref{fig:05}c). The initial ISW1 crestline is 92.3 km long, and the archived vectors give mean speeds of 2.51 and 2.50 m s\(^{-1}\) over 0--4 h and 0--8 h, corresponding to mean displacements of 36.2 and 72.0 km. These cases have comparable crestline lengths but different bathymetric settings. Stripe-geometry extraction preserves their locations, orientations, and curvature, linking SAR target localization to short-term prediction of individual wavefronts.

Stripe segmentation produces binary masks, whereas wavefront construction requires georeferenced vector boundaries, as shown in Fig.~\ref{fig:04}c. Format conversion must therefore preserve the complete stripe area identified by the model. For narrow, curved ISW signatures, contour truncation or incorrect vertex connections can alter the boundary and affect the initial wavefront. To evaluate this conversion, 21 segmentation masks from 12 scenes were vectorized and rasterized back onto the original grid. Spatial overlap before and after conversion was measured by IoU:

\begin{equation}
\label{eq:iou}
\operatorname{IoU}\bigl(M,R(V(M))\bigr)=\frac{\lvert M\cap R(V(M))\rvert}{\lvert M\cup R(V(M))\rvert}.
\end{equation}

Here, M is the original binary mask, V denotes vectorization, and R denotes rasterization onto the original grid. An IoU closer to 1 indicates greater preservation of the stripe area. The original export retained at most 160 sampled vertices, producing 14 invalid polygons and a median IoU of 0.682. Vectorization retaining every foreground pixel produced 21 valid geometries, all with IoU = 1. Full-pixel vectorization thus preserved the original mask extents in the tested samples and provided consistent inputs for wavefront construction. The maximum residual after a pixel-to-geographic coordinate round trip was well below one pixel, supporting numerical consistency of the coordinate transformation.

Candidate selection also depends on model thresholds. Confidence determines which candidates are retained, while the NMS threshold controls suppression of overlapping detections. To assess sensitivity, fixed model weights were used for 108 inference runs across 12 scenes and all combinations of three confidence and three NMS thresholds. At an NMS IoU of 0.30, increasing confidence from 0.07 to 0.25 reduced the total candidate count from 21 to 5 (Table~\ref{tab:03}). Increasing the NMS IoU threshold relaxed overlap suppression and retained more candidates. Thresholds therefore change the target set available for downstream analysis. All other comparisons use confidence 0.07 and NMS IoU 0.30 to control this source of variation.

\begin{table}[!htbp]
\centering
\caption{Total detected instances across 12 scenes under nine threshold combinations.}
\label{tab:03}
\small
\setlength{\tabcolsep}{15pt}
\renewcommand{\arraystretch}{1.18}
\begin{tabular}{@{}lccc@{}}
\toprule
\textbf{Confidence} & \textbf{NMS IoU 0.30} & \textbf{NMS IoU 0.50} & \textbf{NMS IoU 0.70} \\
\midrule
0.07 & 21 & 25 & 36 \\
0.15 & 9 & 11 & 17 \\
0.25 & 5 & 5 & 7 \\
\bottomrule
\end{tabular}
\end{table}

\subsection{Workflow Execution Time}

Per-scene turnaround time was measured for the n8n detection--extraction--prediction chain using all 50 downloaded images. The controller recorded each service interval with a monotonic clock, including initialization, inference, file operations, and failed attempts. Detection and extraction each launched a new process for every call. Total elapsed time extended from the client's n8n request to receipt of a terminal response, excluding language interpretation and online image acquisition.

\begin{figure}[!htbp]
\centering
\includegraphics[width=\linewidth]{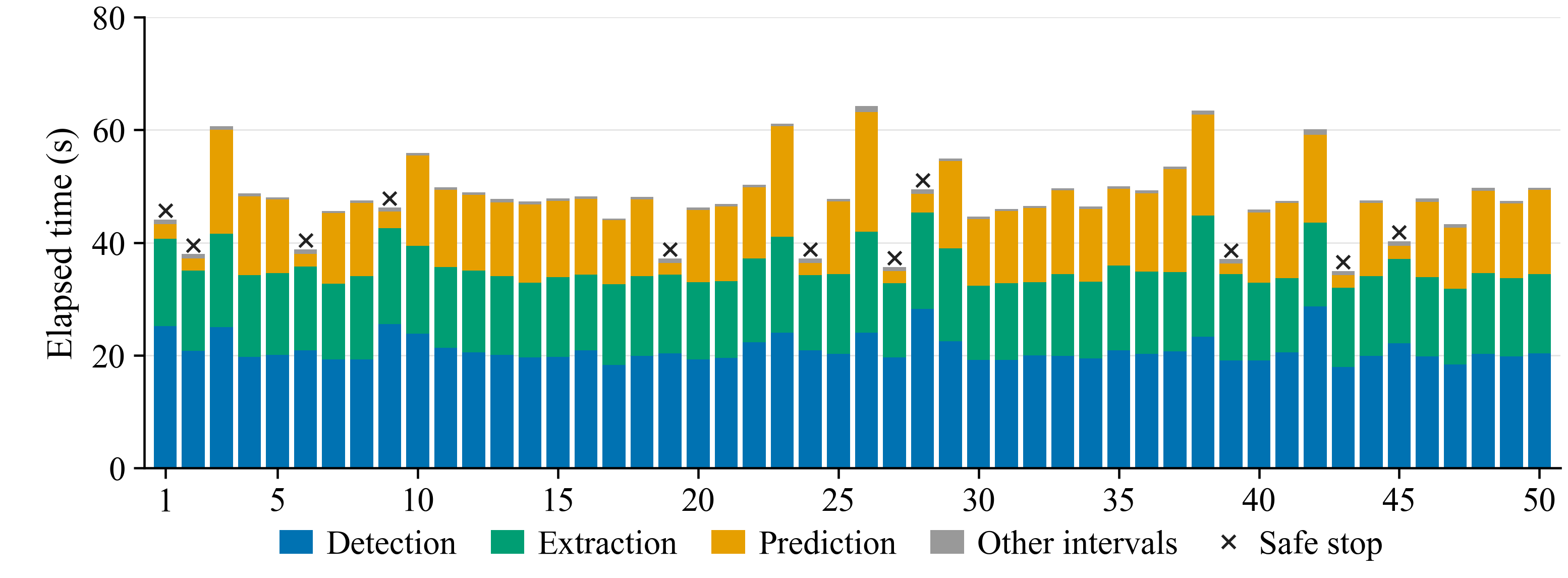}
\caption{n8n execution times for 50 Sentinel-1 scenes. The horizontal axis gives the scene index. Stacked segments show detection, extraction, prediction, and remaining intervals, including failed attempts. Crosses mark the 11 scenes that stopped after empty extraction.}
\label{fig:06}
\end{figure}

Of the 50 scenes, 39 completed detection, extraction, and 3 h prediction with a median turnaround time of 48.04 s and an interquartile range of 47.08--49.92 s (Fig.~\ref{fig:06}). The central 50\% of completed tasks spanned 2.84 s, while the longest took 64.22 s, indicating similar turnaround times for most scenes and a longer tail for a few. Median detection, extraction, and prediction times were 20.10, 14.16, and 13.41 s, respectively. The remaining interval had a median of 0.48 s, with a median share of 0.96\% of total per-scene time. Model services account for most of the time cost, including separate weight loading and image reading for detection and extraction. Service intervals include initialization, inference, and file operations; the remaining interval includes routing and validation. Reducing repeated model loading and image access is therefore a direct route to shorter turnaround times. The other 11 scenes stopped after empty extraction because no wavefront was available, with a median total time of 37.94 s. Fig.~\ref{fig:06} identifies these outcomes explicitly; their shorter durations reflect early termination. The experiments used two Intel Xeon Gold 5118 processors, providing 24 physical cores and 48 logical processors, 160 GiB of RAM, and an NVIDIA RTX 4080 SUPER GPU. Software versions were n8n 2.20.9 and Ultralytics 8.4.77. Detection and propagation used PyTorch 2.12.1+cu132 and 2.1.1+cu118, respectively.

\subsection{Fault Handling and Strategy Ablation}

To separate service recovery from spatial-output validation, 15 fault types and healthy controls were constructed from the actual extraction outputs of 50 scenes, giving 852 paired conditions. Each condition was evaluated with four configurations: the full strategy, no retries, no output validation, and neither mechanism, for 3,408 n8n runs. Evaluation ended at the extraction-service response, with empty feature sets counted as valid in healthy controls. Faults covered transient interruptions, unrecoverable responses, structural defects, wrong-scene outputs, and undeclared feature loss. Independent assessment compared each output with the frozen source product and classified it as valid completion, safe stop, intervention required, or erroneous acceptance. Service faults included HTTP 503, HTTP 429, and timeouts, with recovery after one, two, or three consecutive failures, as well as persistent HTTP 503, HTTP 401 authentication failure, unparseable responses, and application-level errors returned with HTTP 200. Six structural faults were applied to each of 39 scenes: missing files, truncation, self-intersecting geometry, feature-count mismatch, incorrect geometry type, and out-of-range coordinates. Content faults comprised 39 wrong-scene outputs with valid structures and 22 cases in which features were removed and the declared counts adjusted accordingly.

\begin{figure}[!htbp]
\centering
\includegraphics[width=\linewidth]{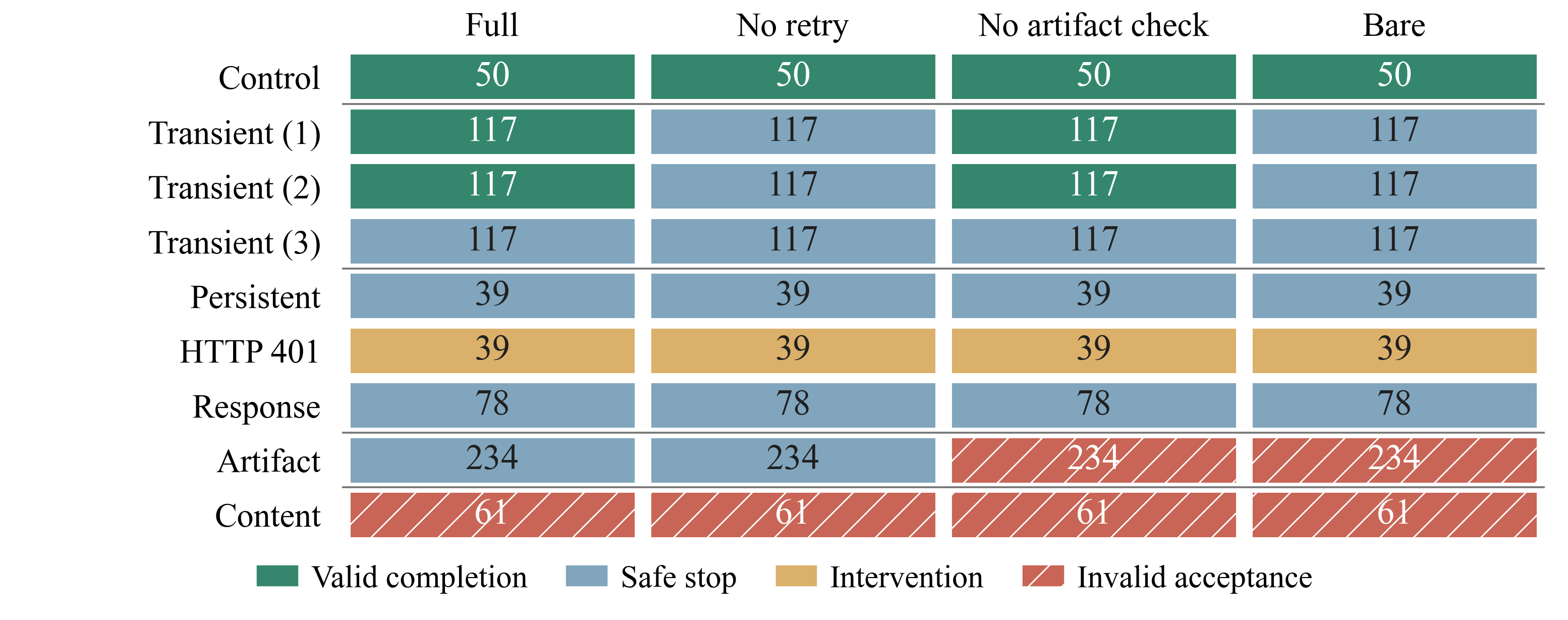}
\caption{Strategy ablation and terminal outcomes under controlled faults. Each column contains the same 852 conditions. Transient(k) combines HTTP 503, HTTP 429, and timeouts, with k denoting consecutive failures. Response comprises two response-error types; Artifact comprises six structural defects; and Content comprises wrong-scene outputs and undeclared feature loss.}
\label{fig:07}
\end{figure}

Retries and output validation address different failure mechanisms (Fig.~\ref{fig:07}). All 234 transient-failure conditions with one or two consecutive failures recovered within the three-attempt budget. Removing retries stopped all these cases, reducing valid completions from 284 to the 50 healthy controls. The full strategy rejected all 234 structural defects. Removing output validation accepted these products and increased erroneous acceptances from 61 to 295. Retries therefore maintained task continuity after temporary interruptions, while validation prevented structurally invalid spatial products from reaching the propagation model. All four configurations missed 61 content errors: 39 wrong-scene products and 22 cases of feature omission with consistent declared counts. These outputs passed file and geometry checks but changed the source or completeness of the ISW crestlines. Cross-model transfer therefore also requires source-image identity checks and verification of the expected feature set. The counts quantify component behavior under the specified fault conditions. Because each scene was reused across conditions, they do not estimate fault incidence during routine operation.

\subsection{Limitations}

This study evaluates task understanding, continuity of spatial processing, and workflow execution in an LLM--n8n ISW processing chain. Scientific accuracy remains dependent on training data and independent observations. Firstly, detection and stripe extraction reuse an existing YOLO11 model; their accuracy and transferability across regions require further evaluation using complete training records and independent regional annotations.  Secondly, the 120 test instructions were generated from predefined rules and provide limited coverage of real-user variation, sustained interaction, and complex task combinations. Independent user tasks are needed to extend the domain instruction set and semantic constraints. Thirdly, workflow performance was evaluated mainly through serial timing and controlled fault tests using Maluku Sea imagery. Performance across regions, concurrent users, and long-term operation requires further quantification.

\section{Conclusion}

ISW-Agent connects LLM-based natural-language understanding, n8n orchestration, and ISW detection, stripe extraction, and propagation-prediction models through a processing chain with explicit input constraints. Task specification identifies the target and computational requirements, geospatial transformations preserve the relationship between images and wavefronts, and stage-specific checks govern data transfer between models. This design links independently configured algorithms into coordinated processing with traceable intermediate products.

Across 120 controlled instructions, GPT-5.6 Sol and DeepSeek correctly identified all operation intents and execution decisions, with a strict task-match rate of 94.2\%. Three archived runs further demonstrated continuous computation from natural-language requests to wavefront prediction for the specified scene. Spatial-fidelity tests showed that full-pixel vectorization preserved the original extents of all 21 model masks. Of 50 timed scenes, 39 completed the full processing chain with a median turnaround time of 48.04 s, while 11 stopped after empty extraction. These results provide complementary evidence for task interpretation, spatial information transfer, and workflow execution.

Fault ablation distinguished service recovery from scientific-output validation. Bounded retries recovered 234 transient interruptions, and structural checks rejected 234 invalid products. The 61 undetected content errors identified the need to verify source-image identity and feature completeness. Reliable dialogue-driven remote sensing therefore requires task continuity, valid spatial structures, and consistent provenance. In one word, ISW-Agent provides a traceable workflow foundation for combining these requirements with independent observational validation and regionally adapted models.

\section*{Acknowledgments and Declarations}

This work was supported in part by the Natural Science Foundation of Shandong Province under Grant ZR2026QC0353 and the National Natural Science Foundation of China under Grant U2006207. The authors acknowledge the n8n platform for providing the workflow automation framework used in ISW-Agent. We thank the European Space Agency (ESA) for providing Sentinel-1 data.

\FloatBarrier
\bibliographystyle{IEEEtran}
\bibliography{references}
\end{document}